\documentclass[runningheads, a4paper, oribibl]{llncs} 
\usepackage{amsmath}
\usepackage{amssymb}
\usepackage{graphicx}
\usepackage{xcolor}
\usepackage{url}

\usepackage{array}

\newcolumntype{C}[1]{>{\centering\let\newline\\\arraybackslash\hspace{0pt}}m{#1}}

\newcommand{\keywords}[1]{\par\addvspace\baselineskip
\noindent\keywordname\enspace\ignorespaces#1}
\begin{document}
\mainmatter  

\title{Ensemble of Hankel Matrices for\\ Face Emotion Recognition} 
\titlerunning{(To appear in ICIAP 2015)}

\author{Liliana Lo~Presti \and Marco La~Cascia}
\institute{DICGIM, Universit\'a degli Studi di Palermo,\\V.le delle Scienze, Ed. 6, 90128 Palermo, Italy,\\ \email{liliana.lopresti@unipa.it}\\DRAFT \\To appear in ICIAP 2015} 
\authorrunning{L. Lo~Presti and M. La~Cascia}

\maketitle

 \begin{abstract} 
In this paper, a face emotion is considered as the result of the composition of multiple concurrent signals, each corresponding to the movements of a specific facial muscle. These concurrent signals are represented by means of a set of multi-scale appearance features that might be correlated with one or more concurrent signals. 
The extraction of these appearance features from a sequence of face images yields to a set of time series. 
This paper proposes to use the dynamics regulating each appearance feature time series to recognize among different face emotions. To this purpose, an ensemble of Hankel matrices corresponding to the extracted time series is used for emotion classification within a framework that combines nearest neighbor and a majority vote schema.
Experimental results on a public available dataset shows that the adopted representation is promising and yields state-of-the-art accuracy in emotion classification.
 \end{abstract}

\section{Introduction} 
\label{sec:Intro}
Emotion recognition deals with the problem of inferring the emotion (i.e. fear, anger, surprise, etc.) given a sequence of face images. Due to strong inter-subject variations, especially in some kind of emotions (such as fear or sadness), and the difficulty to extract reliable feature representations because of illumination changes, biometric differences, and head pose changes, emotion recognition is a challenging problem.
Nonetheless, recognition of face expressions and emotions is of great interest in many fields such as assistive technologies~\cite{LuceyPain},~\cite{Lacava}, socially assistive robotics~\cite{Scassellati}, computational behavioral science~\cite{Rehg},~\cite{Lopresti_Rozga},~\cite{Zeng}, and the emerging field of audience measurement~\cite{Lee}.


A vast literature on affective computing~\cite{Zeng},~\cite{Cavallaro},~\cite{LuceyPain}, has shown that an emotion can be identified by a subset of detected action units. This suggests that face emotion results as combination of movements of various facial muscles. Therefore in this paper we assume that a composition of multiple concurrent signals yields to a face emotion.
We use a restricted set of appearance features -- computed on a frame-per-frame basis -- that may be correlated with one or more of these concurrent signals. 
Given a sequence of face images corresponding to an emotion, the extraction of these appearance features yields to a set of time series, one for each appearance feature. 
Considering that face emotions are not instantaneous, we aim at using the dynamics regulating each sequence of appearance features to recognize among different emotions. 

We propose to model a sequence of face appearance feature as the output of a Linear Time Invariant (LTI) system. Motivated by the success of works in action recognition~\cite{Octavia1},~\cite{ACCV2014}, that represent action-dynamics in terms of Hankel matrices, in this paper we explore the use of Hankel matrices to represent emotion-dynamics. 
We adopt a multi-scale Haar-like feature based appearance representation to obtain a set of time series (one for each spatial scale and Haar-like feature). Hence we represent a sequence of face images by means of an ensemble of Hankel matrices where each Hankel matrix embeds the dynamics of one of the extracted Haar-like feature time series. 
Nearest-Neighbor classifier combined with a majority vote schema is used for classification purposes.

We validated our work on the publicly available extended Cohn-Kanade dataset~\cite{Lucey}. Our experiments show that there is a clear advantage in adopting a dynamics-based emotion representation over using the raw measurements. Furthermore, our experiments highlight that the dynamics of different appearance features contribute differently to the emotion recognition. Overall, our novel emotion representation permits to achieve state-of-the-art accuracy values in comparison to works that use accurate face landmarks.

The plan of the work is as follows. In Section~\ref{sec:RW}, we present works that are related to our emotion-dynamics representation. In Section ~\ref{sec:AR} we describe how we extract a multi-scale face appearance description; Section \ref{sec:ADR} introduces the Hankel matrix-based representation and describes how to build an ensemble of Hankel matrices to describe face appearance dynamics; Section ~\ref{sec:EC} presents details about the adopted classification framework.
Finally, in Sections~\ref{sec:ER} and ~\ref{sec:CFW}, we present experimental results, and conclusions and future directions respectively.


\section{Related Work} 
\label{sec:RW}
Face detection \cite{ViolaJones}, face recognition~\cite{Zhao_Chellappa},~\cite{Lopresti_IMAVIS} and facial expression analysis~\cite{Fasel} have been deeply studied in past years, resulting in a vast literature reviewed in ~\cite{Zeng},~\cite{Cavallaro}. 
In this section, we focus on works that embed the temporal structure of the face image sequence in the feature representation or in the emotion model.

Dynamics-based emotion recognition has been proposed in~\cite{Dibekliouglu} where horizontal and vertical movements of tracked landmarks of different face parts such as eyebrows, eyelids, cheeks, and lip corners jointly with spatio-temporal appearance features are used to describe a sequence of face images. Temporal changes in the face appearance are described by means of  the Complete Local Binary Patterns from Three Orthogonal Planes (LBP-TOP) \cite{Zhao} and classification is performed by SVM. 
While ~\cite{Dibekliouglu} attempts to embed information about the dynamics at a feature representation level, works such as ~\cite{Lorincz},~\cite{Shan_Gong} account for the temporal structure of the sequences of descriptors in the emotion model. 
In~\cite{Nie}, restricted Boltzmann machine with local interactions (LRBM) is used to capture the spatio-temporal patterns in the data. RBM is used as a generative model for data representation instead of feature learning, and data need to be pre-aligned.
In~\cite{Lorincz} time-series kernel methods are used for emotional expression estimation using landmark data only. The work shows that emotion recognition may be done by adopting either the Dynamic Time Warping (DTW) kernel or the Global Alignment (GA) kernel~\cite{Cuturi_GA},~\cite{Cuturi_Vert}.
In ~\cite{Shan_Gong}, a Bayesian approach is used to model dynamic facial expression temporal transitions. Facial appearance representation is computed in terms of Local Binary Patterns (LBP), and an expression manifold is derived for multiple subjects. A Bayesian temporal model (similar to HMM with a non parametric observation model) of the manifold is used to represent facial expression dynamics.

Works such as~\cite{Jeni},~\cite{Wang} use landmarks located on face parts such as eyes, eyebrows, nose and mouth to describe an emotion.
In \cite{Jeni}, a Constrained Local Model (CLM) is used to estimate facial landmarks and extract a sparse representation of corresponding image patches. Emotion classification is performed by least-square SVM. 
Wang et al.~\cite{Wang} propose to use Interval Temporal Bayesian Network (ITBN) to capture the spatial and temporal relations among the primitive facial events. 

Hankel matrices have been already adopted for action recognition in \cite{Octavia1}, which adopts a Hankel matrix-based bag-of-words approach, and in \cite{ACCV2014}, which models an action as a sequence of Hankel matrices and uses a set of HMM trained in a discriminative way to model the switching between LTI systems. In~\cite{AMFG}, we have showed how the dynamics of tracked facial landmarks can be modeled by means of Hankel matrices and can be used for facial expression analysis.

Whilst it is possible to obtain a reasonably accurate estimate of the face region~\cite{ViolaJones}, getting a reliable estimation of facial landmarks is still an open problem despite the remarkable progress described in \cite{Cootes}, \cite{Zhu}. The adoption of appearance feature extracted from the detected face region to describe an emotion, as done indeed in~\cite{Rivera},~\cite{Shan_Gong},~\cite{Zeng},~\cite{Cavallaro}, might be a convenient choice.
Therefore, in this paper we adopt appearance features to represent a face expression. 
In contrast to ~\cite{AMFG}, we do not model landmark trajectories but we use an ensemble of Hankel matrices to describe the dynamics of sequences of appearance features computed at multiple spatial scales. We demonstrate that, without an accurate estimation of facial landmarks, our novel representation can achieve state-of-the-art accuracy in emotion recognition.

\section{Multi-Scale Face Appearance Representation} 
\label{sec:AR}

Given a face image, we need to extract a proper appearance representation for the shown face expression. Considering the success of Haar-like features in face detection we adopt this kind of features to build our face appearance descriptor.

Haar-like features resemble Haar wavelets and have been developed by Viola and Jones for face detection \cite{ViolaJones}. 
A Haar-like feature is computed by considering adjacent rectangular regions in a detection window. The pixel intensities in each region are summed up and the difference between these sums yields the Haar-like feature. In \cite{ViolaJones}, Haar-like features are compared against a threshold and used to detect the face; therefore they are used as weak classifiers and a high number of features are considered in order to build a strong classifier.
The key advantage of a Haar-like feature over most other features is that it can be calculated in constant time due to the use of integral images.

A number of Haar-like features have been used in literature \cite{Lienhart}, \cite{Huang}, and Haar-like features and/or simple variations have been formerly used in literature for emotion recognition \cite{Cavallaro},\cite{YangSim},\cite{Yang} within boosting approaches.

\begin{figure*}[t!]
\centering
\includegraphics[width=0.08\linewidth]{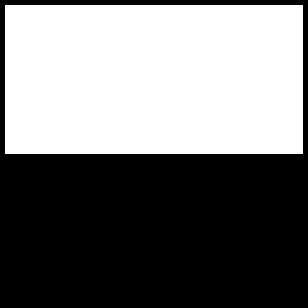}
\includegraphics[width=0.08\linewidth]{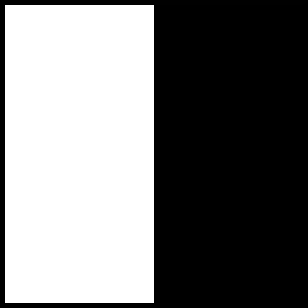}
\includegraphics[width=0.08\linewidth]{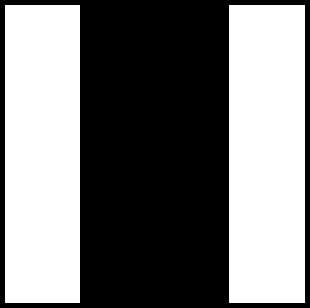}
\includegraphics[width=0.08\linewidth]{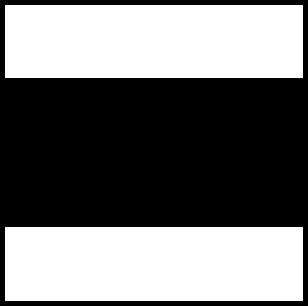}
\includegraphics[width=0.08\linewidth]{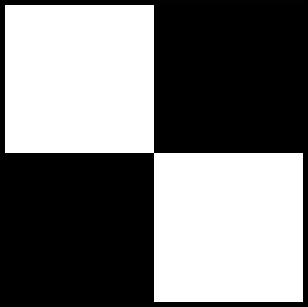}
\includegraphics[width=0.08\linewidth]{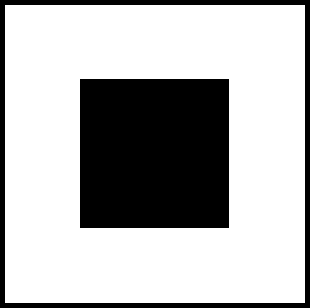}
\vspace{-10pt}
\caption{The set of six Haar-like features used in this paper.}
\label{fig:Haar}
\end{figure*}

\begin{figure*}[t!]
\centering
(a)\includegraphics[width=0.22\linewidth]{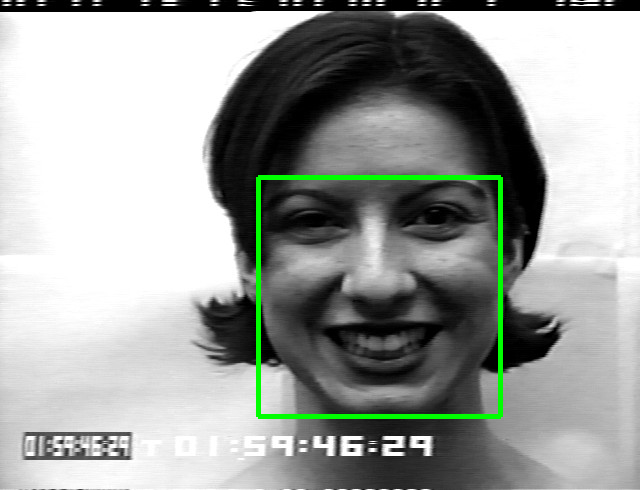}
(b)\includegraphics[width=0.17\linewidth]{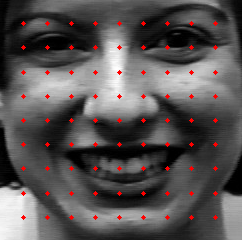}
(c)\includegraphics[width=0.17\linewidth]{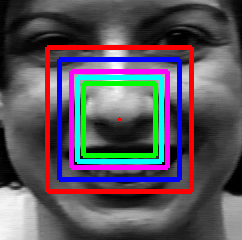}
\vspace{-10pt}
\caption{Haar-like features are extracted from the face region at different spatial scales: (a) the face region is detected and cropped; (b) centers of the sliding window used to compute the Haar-like features; (c) multiple scales used to calculate Haar-like features. }
\label{fig:steps}
\vspace{-10pt}
\end{figure*}

In this paper we only use the six most common features depicted in Figure \ref{fig:Haar}. 
Intuitively, a multi-scale approach might account for different intensity of the emotion, which may change from subject-to-subject. Therefore we extract Haar-like features at different spatial scales.
In this preliminary work, we do not model the weights of each extracted feature; modeling these weights/performing feature selection remains a topic of future investigations.
The main steps we perform to extract our face appearance representation are:
\begin{itemize}
\vspace{-5pt}
\item we detect the face region (as shown in Fig. \ref{fig:steps}~(a));
\item within the face region, we consider a set of uniformly sampled points (red dots in Fig. \ref{fig:steps}~(b));
\item we center windows of varying spatial scales at each of these sampled points (Fig. \ref{fig:steps}~(c) shows the windows centered at a representative point on the subject's nose. Each color indicates a different scale.); 
\item we extract our Haar-like features from each of the selected windows. Whenever the sliding window exceeds the size of the face region (especially along the boundary), the window is cropped so to consider only the pixels within the face area. In our implementation, the white and black rectangular regions of each Haar-like feature are computed in proportion to the window size, therefore the cropping does not affect the computation of the Haar-like features.
\vspace{-10pt}
\end{itemize}

\section{Ensemble of Hankel Matrices for Emotion-Dynamics}
\label{sec:ADR}
In this section, first we briefly review LTI systems and Hankel matrix, then we describe our ensemble of Hankel matrices for emotion-dynamics representation.

\subsection{Hankel Matrix-based Dynamics Representation}
In a LTI system, two linear equations regulate the behavior of the system:
\begin{eqnarray}
&x_{k+1}&=A\cdot x_k + w_k; \nonumber \\
&y_k&=C \cdot x_k.
\label{eq:system}
\end{eqnarray}
The first equation is known as the {\em state equation} and involves the variable $x_{k} \in R^u$, which represents the $u$-dimensional internal state of the LTI system. 
The second equation is known as the {\em measurement equation} and provides a link between the state of the system $x_{k}$ and the $v$-dimensional observable measurement $y_k$.
In such equations the matrices $A$ and $C$ are constant over time, and $w_k \sim N (0, Q)$ is uncorrelated zero mean Gaussian measurement noise.

It is well known \cite{Viberg} that, given a sequence of output measurements $[y_o , \dots, y_\tau]$ from Eq.~\ref{eq:system}, its associated truncated block-Hankel matrix is 
\begin{eqnarray}
\widetilde{H}=\begin{bmatrix}
    &y_0, &y_1, &y_2, &\dots, &y_m \\
    &y_1, &y_2, &y_3, &\dots, &y_{m+1} \\
    &\dots &\dots &\dots &\dots &\dots\\
    &y_n, &y_{n+1}, &y_{n+2}, &\dots, &y_{\tau} \\
   \end{bmatrix},
   \label{eq:hankel}
\end{eqnarray}
where $n$ is the maximal order of the system, $\tau$ is the temporal length of the sequence, and it holds that $\tau=n+m-1$.

The Hankel matrix embeds the observability matrix $\Gamma$ of the system, since $\widetilde{H} = \Gamma \cdot X$, where $X=[x_0, x_1, \cdots, x_\tau]$ is a matrix formed by the sequence of internal states of the LTI system. 

As previously done in ~\cite{Octavia1},~\cite{ACCV2014}, we normalize the Hankel matrix $\widetilde{H}$ as follows:
\begin{equation}
H = \frac{\widetilde{H} }{\sqrt{||\widetilde{H}\cdot \widetilde{H}^T||_F}}.
\end{equation}
and compare two Hankel matrices $H_p$ and $H_q$ by the following similarity score:
\begin{eqnarray}
s(H_p , H_q ) =& ||H_p^T\cdot H_q||_F,
\label{eq:score}
\end{eqnarray}
which can be easily derived from the dissimilarity score in ~\cite{Octavia1}. We have experimentally found that our similarity score is numerically more stable and fast to compute than the dissimilarity score.
Such score can be regarded as an approximation of the cosine of the subspace angle between the spaces spanned by the columns of the Hankel matrices. As such, it can convey the degree to which two Hankel matrices may correspond to the same dynamical system.

\subsection{Emotion-Dynamics Representation}
The simple and fast appearance feature extraction described in Section \ref{sec:AR} yields to a set of time series $Y=\{y^{i,j}\}_{i=1, j=1}^{i=N, j=S}$ where $y^{i,j}=\{y_1^{i,j}, \cdots y_\tau^{i,j}\}$ is the time series corresponding to the $i$-th Haar-like feature at the $j$-th spatial scale ($N$ is the number of Haar-like features, and $S$ is the number of scales). Each element $y_t^{i,j}$ of this time series is a vector of features computed at the uniformly sampled points and representing the $t$-th face in the face image sequence.

We use the set of time series $Y$ to build an ensemble of Hankel matrices $H=\{H^{i,j}\}_{i=1, j=1}^{i=N, j=S}$ where each Hankel matrix $H^{i,j}$ is built upon the time series $y^{i,j}$ and, therefore, is associated with the $i$-th Haar-like feature and the $j$-th spatial scale. Before calculating the Hankel matrix, the sequence $y^{i,j}$ is made zero mean.
We note the following:
\begin{itemize}
\vspace{-5pt}
\item each vector $y_t^{i,j}$ is an ordered set of appearance features extracted from different parts of the face region. The set of Hankel matrices $H^{i,j}$ captures the dynamics of the Haar-like features over the whole face;
\item each Hankel matrix is built upon a single Haar-like feature;
\item each Hankel matrix is built upon a single scale;
\item modeling separately Haar-like features at different spatial scales has computational advantages in terms of memory and time complexity;
\item Hankel matrices can be obtained by a simple and fast reordering of the elements in the vector $y_t^{i,j}$. Therefore, from a computational point of view, the adoption of Hankel matrices over other time series representation is particularly appealing.
\vspace{-5pt}
\end{itemize}

\section{Emotion Classification}
\label{sec:EC}
To test the effectiveness of our novel representation we have adopted the simple and widely used nearest-neighbor classifier (NN). 
We compare Hankel matrices by using the similarity score in Eq. \ref{eq:score}. 
Given an ensemble of Hankel matrices, each Hankel matrix contributes to the emotion classification by voting for a class (predicted by NN). Comparison of Hankel matrices is done on equal terms of Haar-like feature and scale (we compare only Hankel matrices that share the same scale and Haar-like feature). Decision on the predicted class is performed considering a majority vote schema.

Other classification frameworks might be used, such as an LTI system codebook based representation similar to that proposed in \cite{Octavia1}, or a state-based approach similar to that in \cite{ACCV2014}. Alternatively, system identification techniques such as the ones applied in~\cite{Slama},~\cite{Chellappa} can be adopted at the cost of an increased overall time complexity. Even if stronger classification frameworks might be adopted as well, NN allows us to study the effectiveness of our representation without introducing further classifier-dependent parameters.

\begin{table}[h!]
\centering
\begin{tabular}{|c c||c|c|c|c|c|c|c||c|}
\hline
{\bf Features} &{\bf Method}  &{\bf An.} &{\bf Con.} &{\bf Disg.} &{\bf Fear} &{\bf Hap.} &{\bf Sad} &{\bf Surp.} &{\bf Avg}\\
\hline
\raisebox{-.5\height}{\includegraphics[scale=0.05]{haar_like_0.png}} &DTW + NN &37.8 &55.6 &55.9 &16 &73.9 &21.4 &73.5 &47.7\\
\raisebox{-.5\height}{\includegraphics[scale=0.05]{haar_like_1.png}} &DTW + NN &40    &38.9 &32.2 &20 &69.6 &10.7 &54.2 &37.9\\
\raisebox{-.5\height}{\includegraphics[scale=0.05]{haar_like_2.png}} &DTW + NN &40	  &44.4 &22 &20 &63.8 &14.3 &50.6 &36.4\\
\raisebox{-.5\height}{\includegraphics[scale=0.05]{haar_like_3.png}} &DTW + NN &42.2 &66.7 &62.7 &12 &78.3 &10.1 &73.5 &49.4\\
\raisebox{-.5\height}{\includegraphics[scale=0.05]{haar_like_4.png}} &DTW + NN &35.6 &38.9 &54.2 &12 &65.2 &10.7 &66.3 &40.4\\
\raisebox{-.5\height}{\includegraphics[scale=0.05]{haar_like_5.png}} &DTW + NN &57.8 &61.1 &59.3 &16 &68.1 &14.3 &72.3 &49.8\\
\hline
\raisebox{-.5\height}{\includegraphics[scale=0.05]{haar_like_0.png} + \includegraphics[scale=0.05]{haar_like_5.png}} &DTW + NN &53.3 &55.6 &62.7 &16 &72.5 &10.7 &81.9 &50.4\\
\raisebox{-.5\height}{\includegraphics[scale=0.05]{haar_like_1.png} + \includegraphics[scale=0.05]{haar_like_5.png}} &DTW + NN &46.7 &72.2 &52.5 &20 &79.7 &7.1 &67.5 &49.4\\
\raisebox{-.5\height}{\includegraphics[scale=0.05]{haar_like_2.png} + \includegraphics[scale=0.05]{haar_like_5.png}} &DTW + NN &48.6 &55.6 &50.8 &24 &78.3 &17.9 &65.1 &48.6\\
\raisebox{-.5\height}{\includegraphics[scale=0.05]{haar_like_3.png} + \includegraphics[scale=0.05]{haar_like_5.png}} &DTW + NN &60 &55.6 &59.3 &16 &76.8 &14.3 &80.7 &51.8\\
\raisebox{-.5\height}{\includegraphics[scale=0.05]{haar_like_4.png} + \includegraphics[scale=0.05]{haar_like_5.png}} &DTW + NN &53.3 &66.7 &57.6 &20 &78.3 &7.1 &79.5 &51.8\\
\raisebox{-.5\height}{\includegraphics[scale=0.05]{haar_like_2.png} + \includegraphics[scale=0.05]{haar_like_3.png}} &DTW + NN &44.4 &61.1 &50.8 &24 &84.1 &10.7 &73.5 &49.8\\
\hline
{\bf all} &DTW + NN &42.2 &72.2 &59.3 &20 &87 &14.3 &83.1 &54\\
\hline
\hline
\raisebox{-.5\height}{\includegraphics[scale=0.05]{haar_like_0.png}} &Hankel + NN &62.2 &72.2 	  &88.1 &40 &\textcolor{red}{100}  &42.9 &92.8 &71.2\\
\raisebox{-.5\height}{\includegraphics[scale=0.05]{haar_like_1.png}} &Hankel + NN &71.1 &61.1 	  &81.4 &44 &94.2      &64.3 &87.9 &72\\
\raisebox{-.5\height}{\includegraphics[scale=0.05]{haar_like_2.png}} &Hankel + NN &57.8 &61.1 	  &81.4 &44 &97.1 	   &53.6 &84.3 &68.5\\
\raisebox{-.5\height}{\includegraphics[scale=0.05]{haar_like_3.png}} &Hankel + NN &44.4 &66.7 	  &84.7 &40 &{\bf 98.5}   &21.4 &94 &64.3\\
\raisebox{-.5\height}{\includegraphics[scale=0.05]{haar_like_4.png}} &Hankel + NN &77.8 &\textcolor{red}{83.3} &83 &48 &97.1	   &42.9 &90.4 &74.6\\
\raisebox{-.5\height}{\includegraphics[scale=0.05]{haar_like_5.png}} &Hankel + NN &71.1 &{\bf 77.8}		  &91.5 &48 &\textcolor{red}{100}  &60.7 &96.4 &77.9\\
\hline
\raisebox{-.5\height}{\includegraphics[scale=0.05]{haar_like_0.png} + \includegraphics[scale=0.05]{haar_like_5.png}} &Hankel + NN &68.9 &{\bf 77.8} &93.2 &44 &\textcolor{red}{100} &57.1 &96.4 &76.8\\
\raisebox{-.5\height}{\includegraphics[scale=0.05]{haar_like_1.png} + \includegraphics[scale=0.05]{haar_like_5.png}} &Hankel + NN &82.2 &\textcolor{red}{83.3} &91.5 &44 &\textcolor{red}{100} &{\bf 78.6} &91.6 &81.6\\
\raisebox{-.5\height}{\includegraphics[scale=0.05]{haar_like_2.png} + \includegraphics[scale=0.05]{haar_like_5.png}} &Hankel + NN &75.6 &\textcolor{red}{83.3} &89.8 &48 &\textcolor{red}{100} &71.4 &92.8 &80.1\\
\raisebox{-.5\height}{\includegraphics[scale=0.05]{haar_like_3.png} + \includegraphics[scale=0.05]{haar_like_5.png}} &Hankel + NN &60 &72.2 &89.8 &40 &\textcolor{red}{100} &53.6 &94 &72.8\\
\raisebox{-.5\height}{\includegraphics[scale=0.05]{haar_like_4.png} + \includegraphics[scale=0.05]{haar_like_5.png}} &Hankel + NN &{\bf 84.4} &{\bf 77.8} &89.8 &{\bf 56} &\textcolor{red}{100} &64.3 &95.2 &81.1\\
\raisebox{-.5\height}{\includegraphics[scale=0.05]{haar_like_2.png} + \includegraphics[scale=0.05]{haar_like_3.png}} &Hankel + NN &62.2 &{\bf 77.8} &89.8 &44 &\textcolor{red}{100} &57.1 &91.6 &74.6\\
\hline
{\bf all} &Hankel + NN &\textcolor{red}{86.7} &\textcolor{red}{83.3} &\textcolor{red}{96.6} &52 &\textcolor{red}{100} &71.4 &{\bf 97.6} &{\bf 83.9}\\
\hline
\hline
CAPP &SVM~\cite{Lucey} &70 &21.9 &{\bf 94.7} &21.7 &\textcolor{red}{100} &60 &\textcolor{red}{98.7} &66.7\\
LDN  &RBF-SVM~\cite{Rivera}** &71.7 &73.7 &93.4 &\textcolor{red}{ 90.5} &95.8 &\textcolor{red}{78.9} &{\bf 97.6} &\textcolor{red}{85.9} \\
\hline
\hline
Shape (SPTS) &SVM~\cite{Lucey} &35 	&25 		&68.4 	&21.7 	&98.4 	&4 	&100 &50.4\\
Shape+CAPP &SVM~\cite{Chew}	&70.1	&52.4	&92.5	&72.1	&94.2	&45.9	&93.6	&74.4\\
Shape &ITBN~\cite{Wang} &91.1 &78.6 &94 &83.3 &89.8 &76 &91.3 &86.3 \\
Shape  &LRBM~\cite{Nie}	&97.8	&72.2	&89.8	&84		&100	&78.6 	&97.6	&88.6\\
Shape + Hankel &NN ~\cite{AMFG}	&91.1	&83.3	&94.9	&84		&100	&71.4	&98.8	&89.1\\
\hline
\end{tabular}
\caption{Accuracy in Emotion Classification on the CK+ dataset. Red font indicates the best accuracy value per emotion, while bold font highlights the second best performance.  **Different validation protocol (10-fold cross validation)}
\vspace{-15pt}
\label{table:acc}
\end{table}

\begin{table}[h!]
\centering
\begin{tabular}{|C{1.7cm}||C{1.3cm}|C{1.7cm}|C{1.3cm}|C{1.3cm}|C{1.3cm}|C{1.4cm}|C{1.4cm}|}
\hline
{\bf Tr. vs Pr.} &{\bf Angry} &{\bf Contempt} &{\bf Disgust} &{\bf Fear} &{\bf Happy} &{\bf Sadness} &{\bf Surprise}\\
\hline
\hline
{\bf Angry} &{\bf 86.67} &0 &2.22 &2.22 &0 &6.67 &2.22\\
{\bf Contempt} &0 &{\bf 83.33} &0 &0 &5.56 &5.56 &5.56\\
{\bf Disgust} &0 &0 &{\bf 96.61} &0 &1.69 &0 &1.69\\
{\bf Fear} &12 &4 &0 &{\bf 52} &24 &4 &4\\
{\bf Happy} &0 &0 &0 &0 &{\bf 100} &0 &0\\
{\bf Sadness} &7.14 &3.57 &0 &0 &3.57 &{\bf 71.43} &14.29\\
{\bf Surprise} &0 &1.20 &0 &0 &1.20 &0 &{\bf 97.59}\\
\hline
\end{tabular}
\caption{Confusion Matrix on the CK+ dataset when all the six Haar-like features are used. True labels are on rows, and predicted labels are on columns.}
\label{table:CM}
\vspace{-15pt}
\end{table}
\section{Experimental Results}
\label{sec:ER}
We have performed experiments in emotion recognition on the widely adopted Extended Cohn-Kanade dataset (CK+)~\cite{Lucey}. This dataset provides facial expressions of 210 adults. Participants were instructed to perform several facial displays representing either single or combinations of action units. Based on the coded action units and by means of a validation procedure of the assigned label, the segmented recording of the participants' emotions were classified into 7 categories (in brackets the number of available samples): {\it angry (45), contempt (18), disgust (59), fear (25), happy (69), sadness (28), surprise (83)}. 
In total there are 327 sequences of the 7 annotated emotions, performed by 118 different individuals. The number of frames of these sequences ranges in $[6, 71]$ with an average value of about $18 \pm 8.6$.  The dataset provides landmark tracking results obtained by active appearance model, which we use in our experiments only to detect the face region.
We adopted the validation protocol suggested in~\cite{Lucey}, which is leave-one-subject-out cross validation.

When extracting the Haar-like features, we sample the center location uniformly with a step equals to 10\% of the size of the detected face region, yielding a 81 dimensional vector for each Haar-like feature. The spatial scales (size of the window used to calculate the Haar-like feature) are also computed in proportion to the face region size and the percentage ranges in $\{30, 35, 40, 50, 60\}$. The order of each Hankel matrix has been empirically set to 2. 
To extract the Haar-like features we have modified the implementation used in \cite{Hare}, \cite{VISAPP2015}.

\subsection{Results}
\label{ssec:re}

We have performed an extensive validation of the dynamics-based emotion representations whose results are reported in Table \ref{table:acc}. The table reports the per-class classification accuracy values for each of the emotion classes, and the average accuracy.
The table is divided in {\bf 4} parts. The first part presents accuracy values in classification when the raw features are adopted (namely Haar-like features). In this case, as the face image sequences have different lengths, dynamic time warping (DTW) is used to align the sequences and nearest-neighbor classifier is used over the aligned sequences. For a fair comparison, also when adopting the raw features, different Haar-like features and spatial scales are compared separately and a majority vote schema is used to predict the final class.

The second part of the table presents results when an ensemble of Hankel matrices is used.
Both the first and second part of the table report performance when a single Haar-like feature is used, when a pair of Haar-like features is used and, finally, when all the six Haar-like features are used.


By comparing the first and second part of the table, there is a clear advantage in using an ensemble of Hankel matrices to represent the emotions over using directly the Haar-like features. 
On average,  the increase of performance in using the dynamics-based representation with respect to the raw measurements is of about 60.3\%.

Looking at the performance of each single Haar-like feature, the most informative one is the concentric squared regions (the last of the six features). Therefore we have performed experiments to study the performance of this feature when coupled with another Haar-like feature. As the table shows, there is an improvement with three of the five Haar-like features. There is no improvement when the Haar-like feature is coupled with the first Haar-like feature and a degradation of the performance when coupled with the vertical bands Haar-like feature. What is striking is that in all the experiments, the emotion Happy is always correctly recognized 100\% of times. This suggests that our ensemble of Hankel matrices can be appropriate for smile detection.
A further improvement of the performance is obtained when all the Haar-like features are used together, at the cost of an higher computational complexity. We suspect that not all the features are actually contributing to the recognition of the emotion, and feature and scale selection techniques may help to achieve more accurate results.

The third part of the table reports accuracy values of state-of-the-art methods adopting only appearance features. 
The class for which our method seems to fail the most is the emotion {\em Fear}. If we ignore this class, our method achieves even better accuracy values of the most competitive method in \cite{Rivera}. 
For completeness, the fourth part of the table reports the performance of techniques adopting accurate estimation of facial landmarks (provided together with the dataset). Even if these methods are not directly comparable with the ones that use only appearance information, we note that our appearance-based representation competes already very well against these techniques. 

Finally, Table \ref{table:CM} reports the confusion matrix of our method. The class {\em Fear} is confused mostly with the class {\em Happy}. Some confusion is also present between the classes {\em Sadness} and {\em Surprise}.  We believe that these ambiguities might be  probably solved with fine-grained appearance descriptors, such as the Local Directional Number (LDN) pattern introduced in~\cite{Rivera}.

\section{Conclusions and Future Work}
\label{sec:CFW}
In this paper we have proposed to use an ensemble of Hankel matrices to represent the dynamics of face appearance features, where each Hankel matrix embeds the dynamics of a single appearance feature at a given spatial scale. We have tested our novel emotion representation on a widely used publicly available benchmark (CK+).
Our experiments demonstrate that, on equal terms of classification framework and feature representations, the dynamics-based emotion representation achieves about 60.3\% of increase in the accuracy values with respect of using directly the raw measurements. Overall, our approach achieves competitive performance with respect to more sophisticated machinery or methods that use accurate shape information.

Our formulation is general and it is not limited to the adopted face appearance representation. We therefore aim at extending our work by considering other appearance features. Moreover, we believe that feature and scale selection techniques (i.e. boosting) might led to an increase of the accuracy of our approach. 
In this paper, we have focused on the problem of classifying segmented emotion sequences. In future works we aim at tackling with the problem of emotion intensity estimation and emotion detection in face image sequences. In this sense, we will explore how face appearance feature dynamics correlate with the intensity of face emotions and if they can help in detecting subtle changes in face expressions.

\paragraph{\bf Acknowledgments}
This work was partially supported by Italian MIUR grant PON0101687, SINTESYS - Security and INTElligence SYStem.

\bibliographystyle{splncs03}

 \end{document}